\documentclass[runningheads]{llncs}
\usepackage{graphicx}
\usepackage[caption=false, font=normalsize, labelfont=sf, textfont=sf]{subfig}
\usepackage{comment}
\usepackage{amsmath}
\usepackage{amsfonts}

\newcommand{\norm}[1]{\left\lVert#1\right\rVert}
\begin{document}
\title{Lung Structures Enhancement in Chest Radiographs via CT based FCNN Training}
%
%
\author{Ophir Gozes\inst{1} \and
Hayit Greenspan\inst{1}}
\authorrunning{O.Gozes et al.}
\titlerunning{Lung Structures Enhancement in CXR via CT based FCNN Training}
%
\institute{Tel Aviv University, Faculty of Engineering, Department of Biomedical Engineering, Medical Image Processing Laboratory, Tel Aviv 69978, Israel}
\maketitle              
\begin{abstract}
The abundance of overlapping anatomical structures appearing in chest radiographs can reduce the performance of lung pathology detection by automated algorithms (CAD) as well as the human  reader.
In this paper, we present a deep learning based image processing technique for enhancing the contrast of soft lung structures in chest radiographs using Fully Convolutional Neural Networks (FCNN).
Two 2D FCNN architectures were trained to accomplish the task: The first performs 2D lung segmentation which is used for normalization of the lung area. The second FCNN is trained to extract lung structures. To create the training images, we employed Simulated X-Ray or Digitally Reconstructed Radiographs (DRR) derived from 516 scans belonging to the LIDC-IDRI dataset. By first segmenting the lungs in the CT domain, we are able to create a dataset of 2D lung masks to be used for training the segmentation FCNN.
For training the extraction FCNN, we create DRR images of only voxels belonging to the 3D lung segmentation which we call ``Lung X-ray" and use them as target images. 
Once the lung structures are extracted, the original image can be enhanced by fusing the original input x-ray and the synthesized ``Lung X-ray". We show that our enhancement technique is applicable to real x-ray data, and display our results on the recently released NIH Chest X-Ray-14 dataset. We see promising results when training a DenseNet-121 based architecture to work directly on the lung enhanced X-ray images.

\end{abstract}
\section{Introduction}

Chest X-ray is the most frequently performed diagnostic x-ray examination. It produces images of the heart, lungs, airways, blood vessels and the bones of the spine and chest.
It aides in the diagnosis and evaluation of chest diseases such as lung cancer, pneumonia, emphysema, fibrosis, pleural effusion, pneumothorax and tuberculosis.
Lung cancer is the leading cause of cancer death among men and women in the United States and around the world.
In the U.S, according to the American Cancer Society \cite{art0}, in 2018 alone, lung cancer is expected to  account for  25\% of  cancer related deaths, exceeding breast, prostate, colorectal, skin melanoma and bladder cancers combined.

It was found that approx. 90\% of presumed mistakes in pulmonary tumor diagnosis occur in chest
radiography, with only 5\% in CT examinations \cite{art1}.
For this reason, missed lung cancer in chest radiographs is a great source of concern in the radiological community. 

\par
In 2006 Suzuki et al \cite{art4} introduced a method for suppression of ribs in chest radiographs by means of Massive Training Artificial Neural Networks (MTANN). Their work relied on Dual energy X-ray in the creation of training images.
Other recently published works have used Digitally Reconstructed Radiographs (DRR) for training CNN models. Albarqouni et al \cite{art7} used DRR image training for decomposing CXR into several anatomical planes, while Campo et al \cite{art6} used the DRR image training to quantify emphysema severity.
In recent years, with the rapid evolution of the field of deep learning, hand in hand with the release of large datasets \cite{data_ref1,data_ref2}, an opportunity to create a data driven approach for X-ray lung structures enhancement as well as lung pathology CAD has been enabled \cite{art2}.

The current work focuses on enhancement of lung structures in chest X-ray. Our training approach is based on CT data and is focused on extraction of lung tissues and their enhancement in combination with the original radiograph. The enhanced result maintains a close appearance to a regular x-ray, which may be attractive to radiologists. 
The proposed methodology is based on  neural networks trained on synthetic data:
To produce training images, we use  DRRs that we  generated from a subset of LIDC-IDRI dataset.
The LIDC-IDRI dataset (Lung Image Database Consortium image collection) \cite{data_ref1}  consists of diagnostic and lung cancer screening thoracic computed tomography (CT) scans with marked-up annotated lesions. 
\par Given a chest X-ray as input, we introduce a method that allows extraction of lung structures as well as synthesis of an enhanced radiograph.

\vspace{0.1in}
The contribution of this work includes the following:
\begin{itemize}

 \item We present a novel CT based approach to automatically generate lung masks which can be used for training 2D lung segmentation algorithms.
    
\item  We present a novel CT based approach to automatically extract  lung structures in CXR which we term ``lung X-ray" synthesis. The training process for this method makes use of nodule segmentations contained in LIDC-IDRI dataset to introduce a novel nodule weighted reconstruction loss. This assures lungs nodules are not suppressed by the extraction FCNN.
  
\item Combining the above mentioned methods, we present a scheme for lung structures enhancement in real CXR.  
      
\end{itemize}
The proposed method is presented in Section 2. Experimental results are shown in Section 3. In Section 4, a discussion of the results is presented followed by a conclusion of the work.

\section{Methods}

\begin{figure}
\includegraphics[trim=0 700 0 0,clip, width=\textwidth]{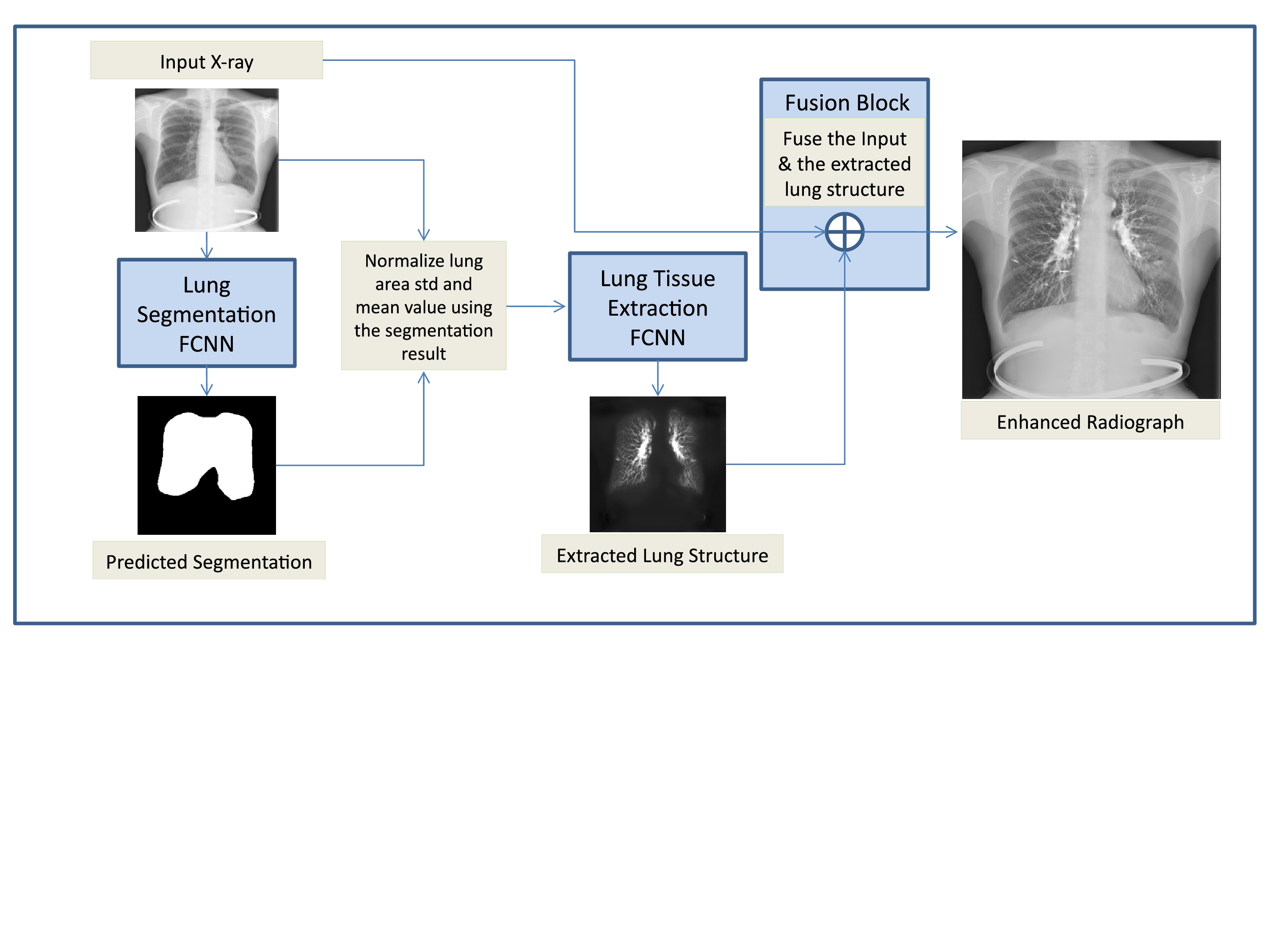}
\caption{A description of the enhancement algorithm  structure.} \label{sysarch}
\end{figure}

 The principal methods presented in the work are the following: 
\begin{description}
\item[$\bullet$ Synthetic chest X-ray generation for training] 
  \item[$\bullet$ 2D segmentation of the lungs:] We present an algorithm for 2D segmentation of the lungs in synthetic X-ray. The algorithm is trained using masks retrieved from CT 3D lung segmentation.
  \item[$\bullet$ Generation of ``lung X-ray"  for training:] Given a CT case as input, we describe a process for the creation of a synthetic X-ray of exclusively the lungs. We term the resulting reconstruction ``lung X-ray" and use it for training.  
  \item[$\bullet$ Lung structures extraction: ] Given a synthetic X-ray as input, a FCNN based algorithm for the synthesis of ``lung X-ray" is presented.
    \item[$\bullet$ Lung structures enhancement:] Combining the above mentioned methods, we present a scheme for lung structures enhancement on real CXR.

\end{description}
Given real chest X-ray as input, the resulting trained lung enhancement method is presented in Fig.\ref{sysarch}. The solution is comprised of a lung segmentation FCNN, a lung structures extraction FCNN, and a fusion block to create the final enhanced radiograph.

\par

\subsection{Synthetic X-ray: Digitally Reconstructed Radiographs}
We begin by introducing the method for generating digitally reconstructed radiographs (DRRs) .\par Our neural networks are trained by using 2D DRRs which serve as input images during the training process. These synthetic X-ray images are generated by reconstructing a three dimensional CT case such that the physical process of 2D X-ray generation is simulated.
In the following subsection we review the physical process which governs chest X-ray generation and review our method for simulating it.
\subsubsection{DRR Generation:}
Our DRR generation method is based on the recently published work by Campo et al \cite{art6}.
As X-rays propagate through matter, their energy decreases.
This attenuation in the energy of the incident X-ray depends on the distance traveled and on the attenuation coefficient.
This relationship is expressed by Beer Lambert's law, where $I_{0}$ is the incident beam, I is the Intensity after traveling a distance x and A is the attenuation coefficient: 
\begin{equation}I = I_{0}\exp^{Ax}
\end{equation}
\par In order to simulate the the X-ray generation process, calculation of the attenuation coefficient is required for each voxel in the CT volume. In a CT volume, each voxel is represented by its Hounsfield unit (HU) value, which is a linear transformation of the original linear attenuation coefficient. Therefore the information regarding the linear attenuation is maintained.
We assume for simplicity a parallel projection model and compute the average attenuation coefficient along the y axis ranging from [1,N] (where N is the pixel length of the posterior anterior view).
Denoting the CT volume by  G(x,y,z), the 2D average attenuation map can be computed using equation 2:
\begin{equation}
 \mu_{av}(x,z) = \sum_{y=1}^{N}  \dfrac{\mu_{water}(G(x,y,z)+1000)}{(N\cdot 1000)}
\end{equation}
Utilizing Beer Lambert's law (Eq 1) the DRR is generated (Eq 3):
\begin{equation}
 I_{DRR}(x,z) = \exp^{\beta \cdot\mu_{av}(x,z)} 
 \label{DRR_create_eq}
 \end{equation}
 The attenuation coefficient of water $\mu_{water}$ was taken as 0.2 $  CM^{-1}$ while $\beta$ was selected as 0.02 such that the simulated X-ray matched the appearance of real X-ray images.

\begin{figure*}[!h]
\centering
\subfloat[]{\includegraphics[height=1.95in]{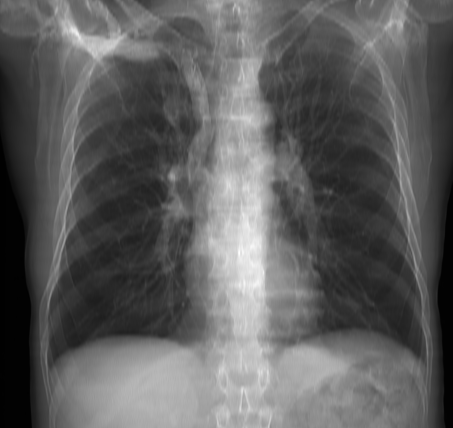}%
\label{fig:lesions_example}}
\hfil
\subfloat[]{\includegraphics[width=2in]{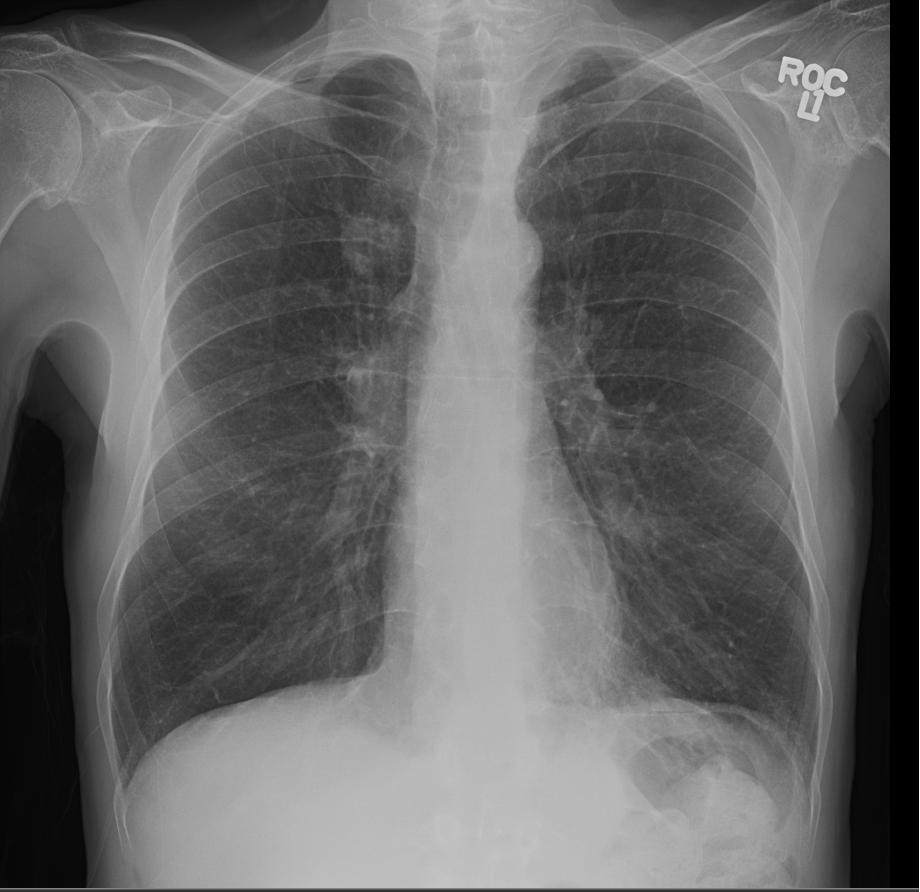}%
\label{fig:roi_extraction_two}}
u\caption{(a) DRR(simulated X-ray) for LIDC-007 (b)  CXR for LIDC-007}
\label{fig:drr_intro}
\end{figure*}

 As some cases in the LIDC-IDRI dataset include radiographs as well as CT data, we were able to compare them to our generated synthetic radiograph. In Fig.\ref{fig:drr_intro}, we present a sample result of the DRR creation process.
The real X-ray and the DRR appear similar in structure. Since our CT dataset contains cases with slice thickness as high as 2.5 mm, the DRR is less detailed then the CXR. 

\subsection{Lung Segmentation in synthetic X-ray}
In the next step of the process, our goal is to segment the lung region in a given synthetic X-ray (DRR). For this we employ a FCNN and train it as detailed next.

\subsubsection*{Creation of 2D lungs masks for training: }
For each synthetic X-ray, a 2D training mask can be generated by a 2D projection of the 3D lung mask matching the CT case used for DRR generation.
In order to  create a 3D lung mask of the lung volume in CT,  we first perform binarization of the CT scan $G(x,y,z)$ with a threshold of -500[HU]. For each axial slice we extract the filled structure of the largest connected component. The 2D axial segmentations are then stacked to create a 3D binary mask  $M_{lung}3D$ of the entire volume.
In order to create 2D masks to accompany the 2D DRR's we project the binary $M_{lung}3D$ along the y axis, yielding a 2D mask which we denote 
$M_{lung}2D$.
\par
\begin{figure*}[!h]
\centering

\subfloat[]{\includegraphics[height=2in]{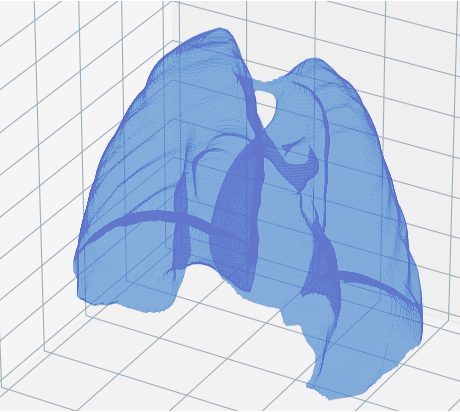}%
 \label{fig:lesions_example}}
\hfil
\subfloat[]{\includegraphics[height=2in]{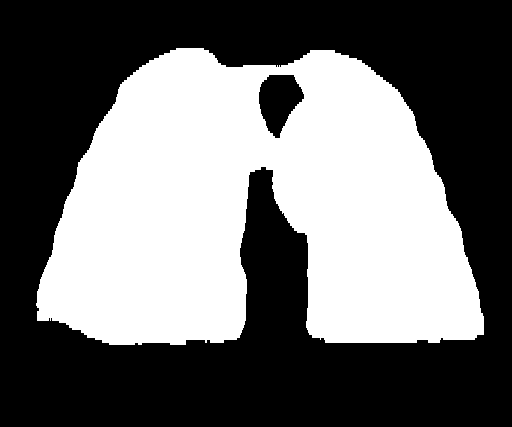}%
\label{fig:roi_extraction_two}}
\caption{(a) 3D Lung segmentation Mask (b) 2D Projection of the Lung Mask}
\label{fig:Lung Segmentation Gt}
\end{figure*}

In contrast to 2D masks usually employed in the process of training 2D lung field segmentation algorithms \cite{art5}, the masks generated by our  method reflect the exact position the of lungs in the image even when occluded by other structures. As a result, the subdiaphagramatic and retrocardiac areas  which are known as hidden spots for nodule detection \cite{art1} are included in the mask.
An example of a 3D mask and a 2D mask is given in Fig. \ref{fig:Lung Segmentation Gt}

\subsubsection{Segmentation FCNN: }
The networks we use in this work are based on the U-net FCNN architecture \cite{art3}. 
We specify here the modifications which we made to the original architecture:
The inputs size to the segmentation network is $512\times512$ with 32 filters in the first convolution layer. In order to improve generalization of the network, we add a Gaussian noise layer(std=0.2) which operates on the input .
We use dilated convolutions (dilation rate = 2) in order to enlarge the receptive field of the network .
Batch normalization layers were not used.
For nonlinearity, RELU activation is used throughout the net, while at the network output we use a sigmoid activation.
The output size is $512\times512$.
The loss function we used is weighted binary cross entropy.
\par
Training was performed on batches of size 8. ADAM optimizer was used. The optimal initial learning rate was found to be 1E-4. Validation loss converged after 80 epochs.
To augment the dataset during the training we utilized random data augmentation for both the source and the lung mask. We used random rotations by 2 degrees, width shift with range 0.1, height shift with range 0.2, random zooming with factor 0.3.

\subsection{Lung Structures Extraction Method}
The input to the lung structures extraction algorithm is a synthetic X-ray. The output is an image which includes only the lung structures appearing in the original image. In order to teach a FCNN to perform this decomposition task,  
we make use of the 3D CT data. A DRR of the lungs which we term ``Lung X-ray" is created and used as the training target image.
Training pairs of source and target images are generated as detailed next: \subsubsection {Source ``Synthetic X-ray'' DRR image Generation:} For each CT case we produce a DRR (Eq.\ref{DRR_create_eq}) which serves as a ``Synthetic X-ray'' source image.
\subsubsection {Target ``Lung X-ray'' image Generation:}

\begin{figure*}[!h]
\centering
\subfloat[]{\includegraphics[height=2in]{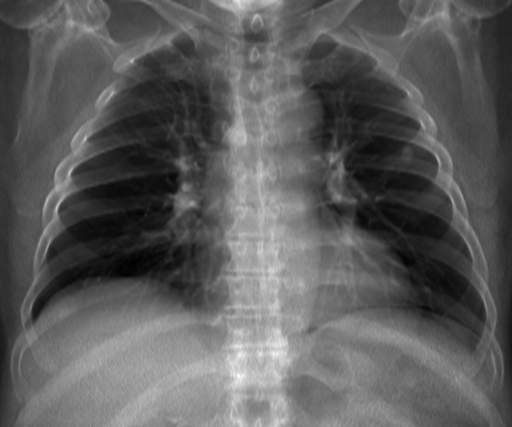}%
\label{fig:lesions_example}}
\hfil
\subfloat[]{\includegraphics[height=2in]{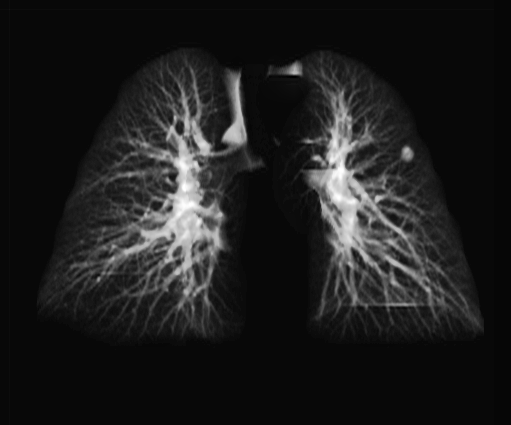}%
\label{fig:roi_extraction_two}}
\caption{(a) DRR Source Image:  (b)  ``Lung X-ray" Target Image for Training }
\label{fig:DRR and Target}
\end{figure*}

Utilizing the 3D segmentation map $M_{lung}$, we mask out all non Lung voxels yielding $G_{Lung}$.
\begin{equation}
G_{Lung}=M_{lung}\cdot G(x,y,z)
\end{equation}
A DRR can now be generated as before for $G_{Lung}$ using Eq.3. As a consequence, the DRR generation process is now limited to the lung area. 
An example result of the ``Lung X-ray'' generation process is given in Fig.\ref{fig:DRR and Target}. It is noticeable that only inner lung structures appear, excluding overlapping anatomical structures.


\subsubsection{Lung Structures Extraction FCNN:}
By training a FCNN to synthesize a ``Lung X-ray'' matching a DRR input, we are able to extract the lung structures from the input image.

In the following we describe the FCNN used in the lung structures extraction algorithm. 
We used the U-Net Architecture \cite{art3} with the following modifications: 
The inputs size is $512\times512$ with 32 filters in the first convolution layer.
We use RELU activation functions throughout the net while at the network output we use the Tanh activation.
The output size is $512\times512$.
\par
In order to ensure small structures such as nodules appear in the  synthesized ``Lung X-ray'', we assigned higher loss weight to image areas which contained nodules.
For each DRR, a matching 2D binary mask of nodules was generated by projecting the 3D CT nodule annotations which are available in the LIDC dataset. \par
The resulting loss function (Eq.\ref{nodule_loss})  is a weighted L1 loss computed between the FCNN output $y_{pred}$ and the GT target ``Lung X-ray'' image.

\begin{equation}
L1_{weighted}=\norm{(y_{pred} - target)\cdot(1+ w_{noduleLoss}\cdot noduleMask)}_{1}
\label{nodule_loss}
\end{equation}

Nodules with the following features were selected:
median texture greater then 3,  median subtlety greater then 4, level of agreement of at least 2  radiologists. This was performed in order to ensure that the nodules used are actually visible in the DRR .
A suitable $w_{noduleLoss}$ was found to be 30.
\subsubsection{Preprocessing  }

Images have been normalized to be in the range of [0,1] and have been equalized using by first HE, then CLAHE with window of [40,40] and contrast clip limit 0.01.
When working with real X-ray, before feeding the images to the extraction network, we use the segmentation FCNN to segment the lungs and then normalize the lung area to mean 0, and std 0.5.
     
\subsubsection{Training of the Lung Extraction FCNN }
The network was trained to synthesize the required target image using training with batches of size 8 and with ADAM optimizer. The optimal initial learning rate was found to be  1E-4. Validation loss converged after 200 epochs.

To augment the dataset during the training we utilized random data augmentation for both the source and the target. We used random rotations by 4 degrees, width and height shift with range 0.1, random zooming with factor 0.2, and horizontal flipping. 

\subsection{A Scheme for Lung Structures Enhancement in Real CXR}
Once extracted, the lung structures can be added to the original DRR image, allowing for a selective enhancement of lung structures.
For the enhancement of real CXR (Fig.\ref{sysarch}), we first segment the lung area using the segmentation FCNN. We proceed by normalizing the lung area to mean 0, std 0.5.
Following the normalization procedure, we use the lung structures extraction FCNN to extract lung structures from the input image (i.e prediction of a ``Lung X-ray").
The input CXR image and the synthesized ``Lung X-ray" are scaled to the range of [0,1]. We fuse the two images by performing a weighted summation (Eq.\ref{Enhancement_Eq}).
\begin{equation}
I_{Enhanced}=I_{CXR} + w\cdot I_{LungXray} 
\label{Enhancement_Eq}
\end{equation}

\begin{figure}
\centering\includegraphics[trim=0 700 1200 0,clip, width=\textwidth]{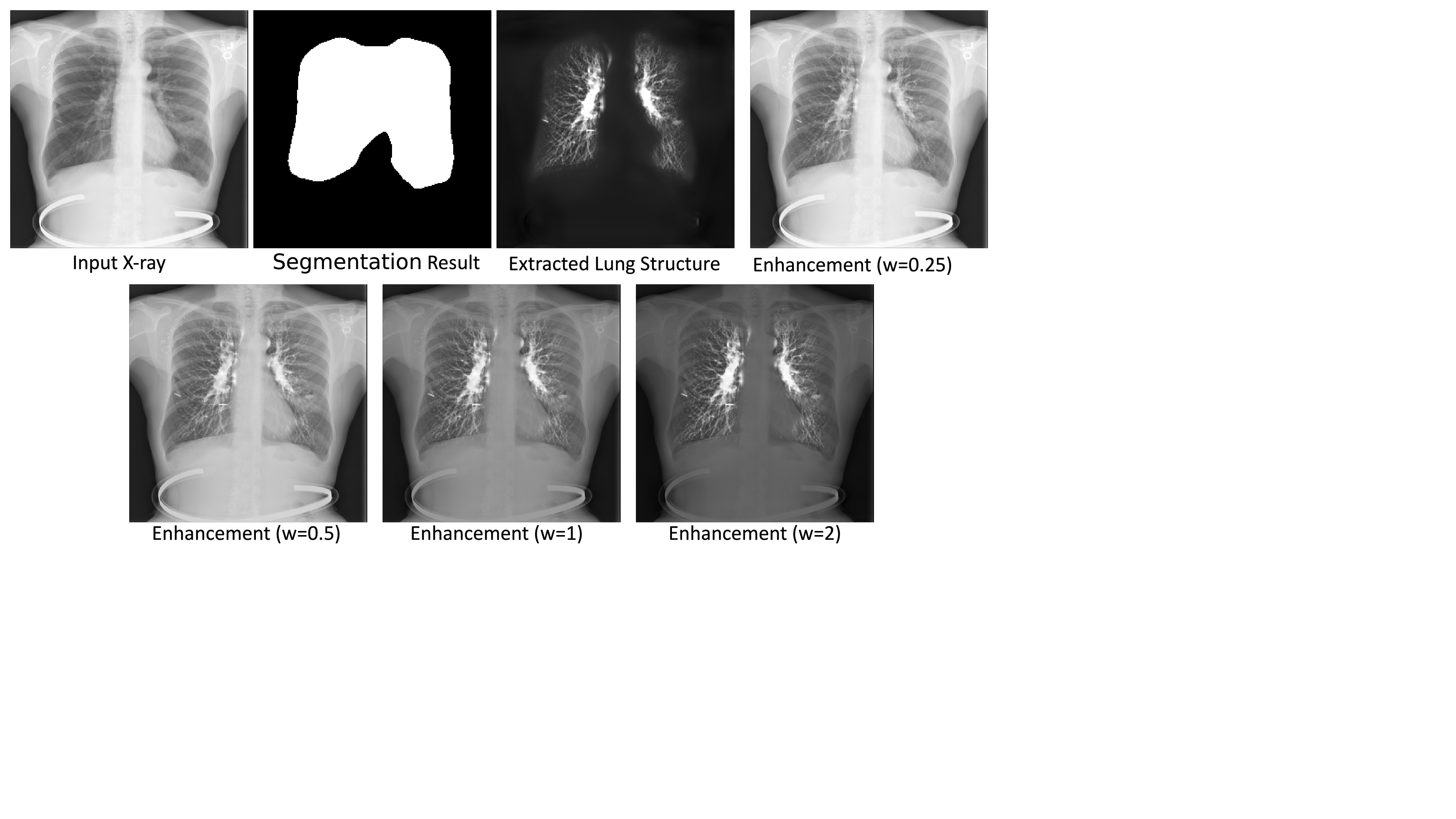}
\caption{NIH Chest X-Ray-14 case \#1555 Enhancement Results} \label{enhancment2rows}
\end{figure}

In Figure \ref{enhancment2rows} we display example results on a real chest X-ray image. An enhancement weight factor $w$ is used to factor the extracted lung image.  By controlling $w$, multiple enhancement levels can be achieved.

\section{Experiments and Results}
\subsection{Lung Segmentation}

\begin{figure*}

\centering

\subfloat[]{\includegraphics[height=1.6in]{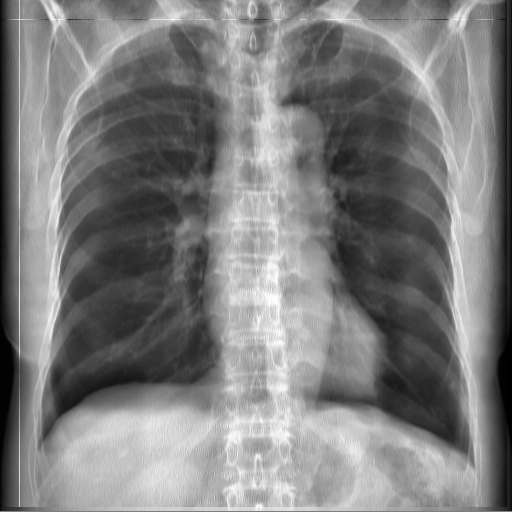}%
\label{fig:DRR input}}
\hfil
\subfloat[]{\includegraphics[height=1.6in]{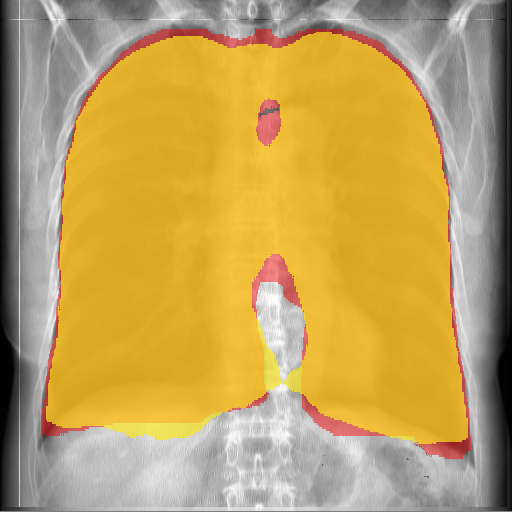}%
\label{fig:segmentation output}}
\caption{(a) Original DRR  (b) Segmentation Result (Red: GT Mask derived from CT, Yellow: Segmentation result, Orange: overlap)}
\label{segmentation results}
\end{figure*}

A total of 990 CT cases were used to create pairs of DRR and Lung mask for training the segmentation FCNN. Data was split in 80-20 training validation ratio.
We threshold the continuous output of the segmentation network to get a binary segmentation prediction (threshold=0.95).
An example result of the segmentation algorithm is given in Figure \ref{segmentation results} .
The Dice coefficient was chosen as the segmentation metric.On the training set the Dice score was \textbf{0.971} while on the validation set the Dice score was \textbf{0.953}.
Since our lung segmentation algorithm is used for normalization of the input to the extraction FCNN we were satisfied with the result.
 To the best of our knowledge, this is the first work that performs lung segmentation using CT ground truth. As a result, we were not able to perform a comparison to other works.

\subsection{Lung Structures Extraction }

\begin{figure*}[!h]
\centering
\subfloat[]{\includegraphics[height=1.4in]{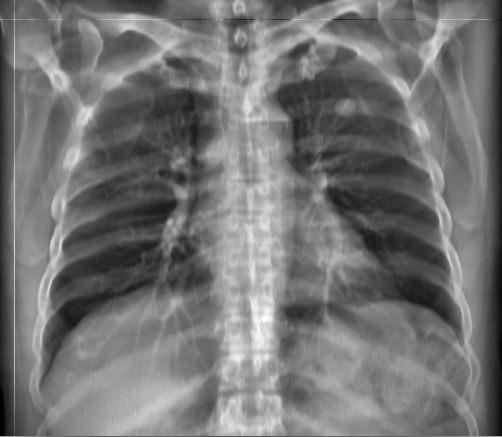}%
\label{fig:DRR input}}
\subfloat[]{\includegraphics[height=1.4in]{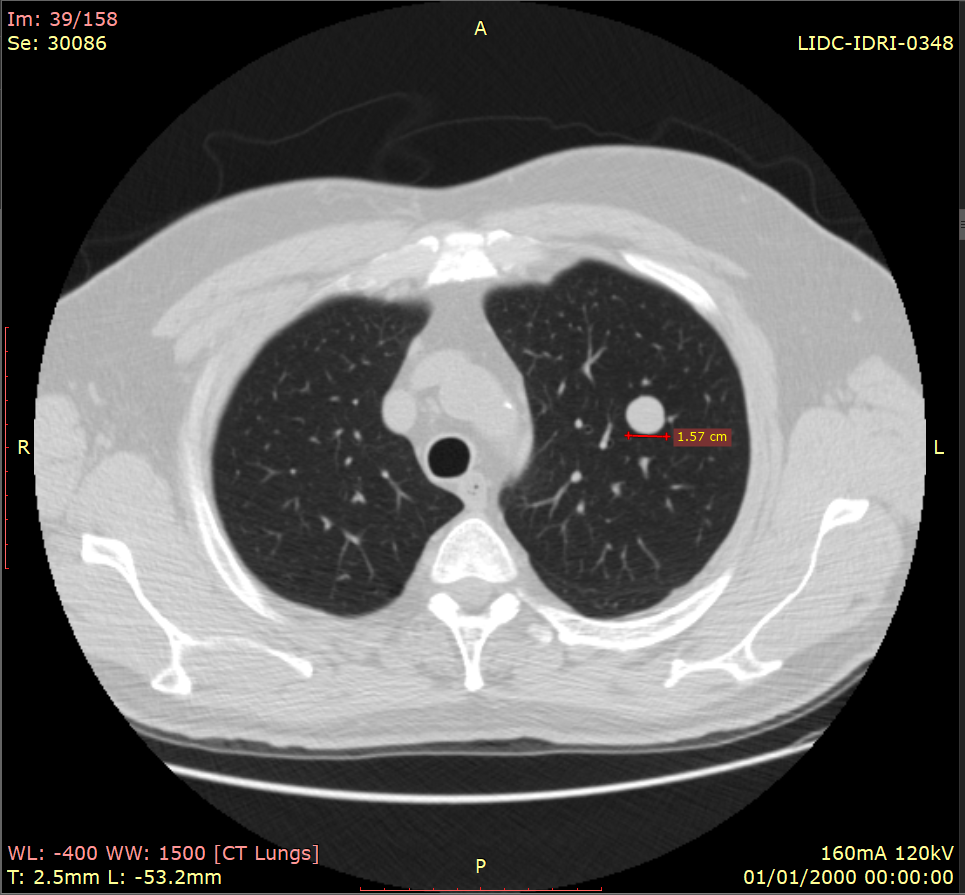}%
\label{fig:segmentation output}}
\subfloat[]{\includegraphics[height=1.4in]{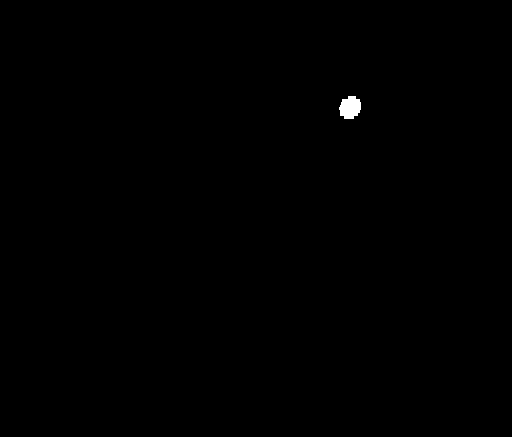}%
\label{fig:segmentation output}}\vfill  \vspace{-4pt}
\subfloat[]{\includegraphics[height=1.4in]{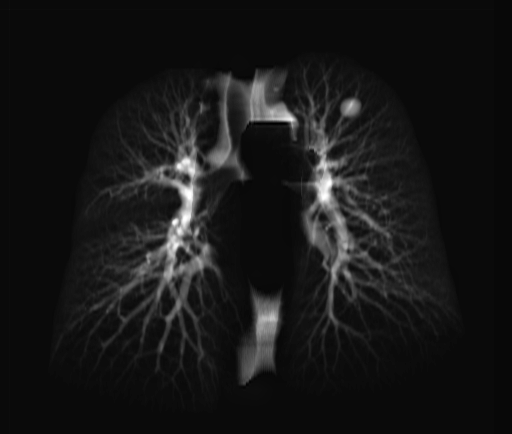}%
\label{fig:segmentation output}}
\subfloat[]{\includegraphics[height=1.4in]{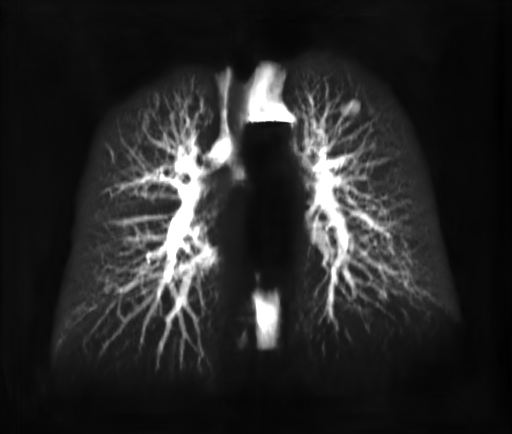}%
\label{fig:segmentation output}}
\subfloat[]{\includegraphics[height=1.4in]{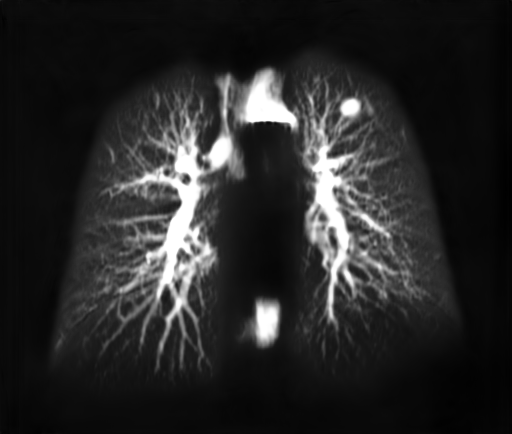}%
\label{fig:ext_out_30}}

\caption{(a)  DRR  (b) CT view of a solid 1.6CM nodule (c)Nodules mask created by projecting LIDC annotations to 2D  (d) Target GT- ``Lung X-ray" (e)  Synthesized result  without nodule weighted L1 loss (f)  Synthesized result  ($w_{noduleLoss}=30$)  }
\label{Extraction results}
\end{figure*}

A subset of 516 CT scans belonging to LIDC dataset was used for training.
For each CT case we generate a DRR and a ``Lung X-ray" pair which we denote by $I_{Source}$ and $I_{Target}$.
In addition, we generate a 2D binary mask of the nodules that belong to the case and use it for the computation of the loss function(Eq.\ref{nodule_loss}). 
We split the dataset to 465 training pairs and 51 validation pairs.
We evaluate our results on 51 validation cases and report MAE (Mean Absolute Error), MSE, PSNR, and SSIM.
Results are given in Table.\ref{tab2}.
An example result of the extraction algorithm is given in Figure \ref{Extraction results}.
In this case, the CT (\ref{Extraction results}.b) contained one nodule which was projected to create the nodules mask (\ref{Extraction results}.c).
Notice that the introduction of our weighted nodule loss function greatly improves the visibility of the nodule in the extracted result(\ref{Extraction results}.e vs \ref{Extraction results}.f)

 \begin{table}[!h]
 \centering
\caption{Extraction Network Performance Results }\label{tab2}
\begin{tabular}{|c|c|c|c|c|}
\hline
 &  MAE & MSE & PSNR [dB] &SSIM\\
\hline
    {Average} &    \textbf{0.082} &   \textbf{0.03} &
\textbf{24.98} &   \textbf{0.80}\\
\hline
    {Std} &    \textbf{0.017} &   \textbf{0.007} &
\textbf{1.41} &   \textbf{0.04}\\
\hline
\end{tabular}
\end{table}
\vspace{-40pt}

\subsection{Applicability to Real X-ray}
In order to explore the applicability of our algorithm to real X-ray and  to examine whether the enhancement scheme introduces artifacts detrimental to CAD detection performance, we chose to perform the lung enhancement algorithm as a preprocessing step on nodule and mass CAD input images.

\begin{figure*}[!h]
\centering
\subfloat[]{\includegraphics[height=1.4in]{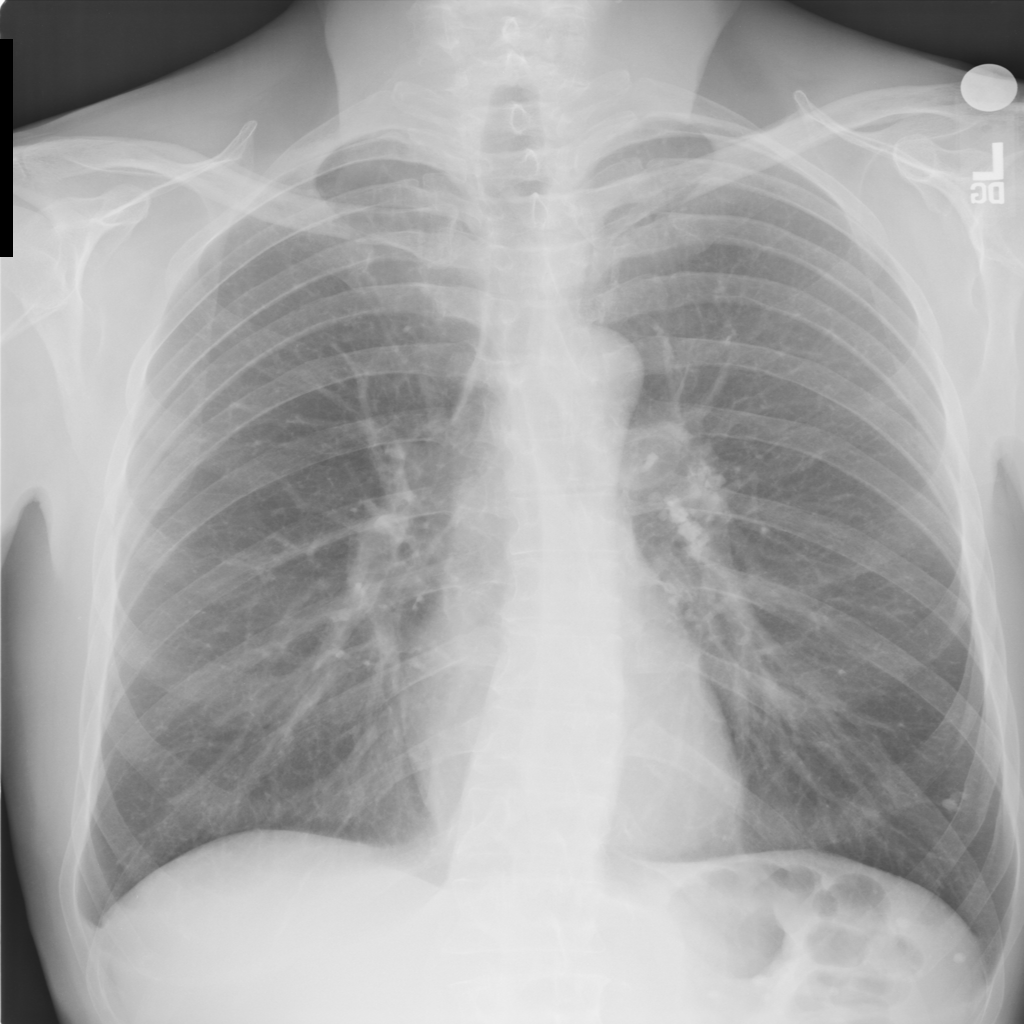}%
\label{fig:lesions_example}}
\hfil
\subfloat[]{\includegraphics[height=1.4in]{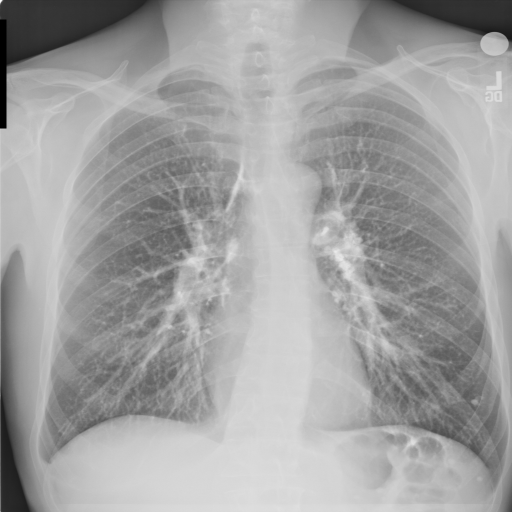}%
}

\label{fig:use_case_en}
\centering
\subfloat[]{\includegraphics[height=1.4in]{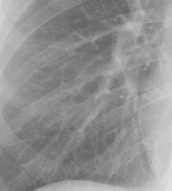}%
\label{fig:lesions_example}}
\hfil
\subfloat[]{\includegraphics[height=1.4in]{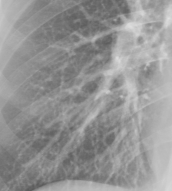}%
}

\label{fig:use_case_en}
\caption{Enhancement result on NIH ChestX-Ray14 image. One network was trained on the original X-ray and a second network was trained on the enhanced X-ray (a)  Original X-ray (b) Enhanced X-ray (c) Zoom (d) Zoom -Enhanced }
\label{fig:enhancement_on_xray}
\end{figure*}

To accomplish this, we trained and tested a CheXNet \cite{art2} based network on 67,313 images released in ChestX-ray14 dataset(subset of PA images).
In Figure \ref{fig:enhancement_on_xray} we show the results of the enhancement on an image from NIH ChestX-ray14 dataset \cite{data_ref2}.
The architecture we chose was a DenseNet-121 based network, with $512\times512$ input size. Network weights were initialized with pretrained ImageNet weights and training was performed independently for enhanced images and non-enhanced images.

The dataset was split to 44,971 training, 11,245 validation and 11,097 test images. Results are given on the official test set in terms of average precision (AP) for the labels mass and nodule. We see a moderate increase in AP scores for the mass detection task (Table  \ref{tab3}).

 \begin{table}[!h]
 \centering
\caption{Effect of lung structure enhancement preprocessing on CAD AP\%. Evaluated using 5000 bootstrap replicates, given as mean(std).  }\label{tab3}
\begin{tabular}{|c|c|c|}
\hline
  &
\textbf{Mass} &
\textbf{Nodule} 
\\\hline
 Non Enhanced   &
 36.83(1.58) &
29.25(1.53)
\\\hline Enhanced &
 \textbf{39.65(1.74)} &
 \textbf{29.73(1.56)}\\\hline

\end{tabular}
\end{table}
\vspace{-30pt}

\section{Discussion and Conclusion}
In this work we  presented a novel method for enhancement of lung structures in chest X-ray.
We demonstrated that 3D CT data, along with 3D nodule annotations, can be used to train a 2D lung structures enhancement algorithm.
By generating our own synthetic data, we enable neural networks to train on ground truth images which are not achievable by current X-ray techniques.
Initial results suggest a moderate improvement of lung mass CAD can be achieved by performing the proposed lung enhancement scheme as a preprocessing step. The results also indicate that the effect of artifacts that may have been introduced by the enhancement scheme is minimal.
We plan to study next the impact that the enhancement algorithm can have on a human reader performance.
In future work, we plan to improve the robustness of our method by performing unsupervised domain adaptation between the CXR domain and the synthetic DRR domain.

\subsubsection{Acknowledgements: } We thank Prof. Edith Marom from Sheba Medical Center for providing clinical consultation. We appreciate valuable suggestions from Avi Ben-Cohen in Tel-Aviv University.


\end{document}